\pdfoutput=1
\documentclass[conference]{IEEEtran}
\IEEEoverridecommandlockouts
\usepackage{cite}
\usepackage{amsmath,amssymb,amsfonts}
\usepackage{algorithmic}
\usepackage{graphicx}
\usepackage{textcomp}
\usepackage{xcolor}
\def\BibTeX{{\rm B\kern-.05em{\sc i\kern-.025em b}\kern-.08em
    T\kern-.1667em\lower.7ex\hbox{E}\kern-.125emX}}

\usepackage{color, colortbl}
\usepackage{bm}
\usepackage[export]{adjustbox}
\usepackage{svg}
\usepackage{multirow}
\usepackage{makecell}
\usepackage{booktabs}
\usepackage{url}
\usepackage[hidelinks]{hyperref}
\hypersetup{
    colorlinks=true,
    linkcolor=black,
    filecolor=black,
    urlcolor=gray,
    citecolor=black,
    breaklinks=true
}

\usepackage{tikz}
\usepackage{pgfplots}
\pgfplotsset{width=7cm,compat=1.16}
\usepgfplotslibrary{statistics}
\usetikzlibrary{pgfplots.groupplots}
\usetikzlibrary{plotmarks}
\usetikzlibrary{shapes}

\def\x{{\bm{x}}}

\definecolor{c1}{RGB}{ 246.2025  131.8860    8.7975}    % SetDesc
\definecolor{c2}{RGB}{   128.0100  128.0100  128.0100}  % LMED
\definecolor{c3}{RGB}{          0  202.2405  255.0000}  % TD
\definecolor{c4}{RGB}{          0         0  255.0000}  % TDS
\definecolor{c5}{RGB}{          0  255.0000         0}  % MST-S
\definecolor{c6}{RGB}{          0   87.9240         0}  % MST
\definecolor{c7}{RGB}{          0   0         0}  % BLACK
\definecolor{Gray}{gray}{0.9}

\makeatletter
\newcommand{\linebreakand}{%
  \end{@IEEEauthorhalign}
  \hfill\mbox{}\par
  \mbox{}\hfill\begin{@IEEEauthorhalign}
}
\makeatother

\begin{document}

\title{Data augmentation with mixtures of max-entropy transformations for filling-level classification}

%\title{Improving filling level estimation with principled data augmentation}

\author{\IEEEauthorblockN{Apostolos Modas}
\IEEEauthorblockA{\textit{EPFL}\\
Switzerland\\
apostolos.modas@epfl.ch
}
\and
\IEEEauthorblockN{Andrea Cavallaro}
\IEEEauthorblockA{\textit{Queen Mary University of London}\\
United Kingdom\\
a.cavallaro@qmul.ac.uk}
% \linebreakand
\and
\IEEEauthorblockN{Pascal Frossard}
\IEEEauthorblockA{\textit{EPFL}\\
Switzerland\\
pascal.frossard@epfl.ch
}
\thanks{\textcopyright 2022 IEEE. Personal use of this material is permitted. Permission from IEEE must be obtained for all other uses, in any current or future media, including reprinting/republishing this material for advertising or promotional purposes, creating new collective works, for resale or redistribution to servers or lists, or reuse of any copyrighted component of this work in other works.}
}

\maketitle

\begin{abstract}
We address the problem of distribution shifts in test-time data with a principled data augmentation scheme for the task of content-level classification. In such a task, properties such as shape or transparency of test-time  containers (cup or drinking glass) may differ from those represented in the training data. Dealing with such distribution shifts using standard augmentation schemes is challenging and transforming the training images to cover the properties of the test-time instances requires sophisticated image manipulations. We therefore generate diverse augmentations using a family of max-entropy transformations that create samples with new shapes, colors and spectral characteristics. We show that such a principled augmentation scheme, alone, can  replace current approaches that use transfer learning or can be used in combination with  transfer learning to improve its performance.
\end{abstract}


%%%%%%%%%%%%%%%%%%%%%%%%%%%%%%%%%%%%%
\section{Introduction}
\label{sec:intro}

In many machine learning tasks, transfer learning~\cite{TanTransferLearning} helps overcome the limited amount of training data~\cite{huh2016makes,Kornblith_2019_CVPR,XueTexture,Farhadi2009,Sajjan2020ICRA_ClearGrasp,Vedaldi2014CVPR}. However, pre-training models on large datasets designed for another task gives no guarantees on the relevance of  the transferred features for the target task. An alternative to increase training-data variability is with data augmentation~\cite{Shorten2019SurveyDataAugm}. However, generating new training samples that are representative of test-time data might require complex operations, including composition of transformations or mixing strategies~\cite{Hendrycks2020AugMix}.

Robot perception offers several examples of real-world machine learning tasks that rely on scarce training data. One of such examples is human-robot interaction, where estimating the physical properties of objects is important for executing safe grasps~\cite{Sanchez-Matilla2020}. An important property to be estimated is the weight of the object (e.g.~a container with unknown content) through computer vision, which requires the inference of the shape of the container~\cite{Xompero2020ICASSP_LoDE}, and the type and amount of its content~\cite{Mottaghi2017ICCV,Modas2021Improving}. Classifying the filling level of an \emph{unknown} container, though, is challenging due to test-time distribution shifts caused by hand occlusions, transparencies of both the container and the filling~\cite{Sajjan2020ICRA_ClearGrasp}, and shape differences among containers. Current approaches  use RGB~\cite{Mottaghi2017ICCV}, thermal~\cite{Schenck2017ICRA}, or a combination of RGB and depth data~\cite{Do2016,Do2018} and cannot rely on large task-specific  training data. Therefore, transfer learning with classifiers pre-trained on large datasets (i.e.~ImageNet~\cite{Deng2009CVPR_ImageNet}) is typically used~\cite{Mottaghi2017ICCV,Modas2021Improving} instead of data augmentation. Although data augmentation imposes explicitly the changes that the classifiers should be invariant to (e.g.~random noise, random horizontal flips), typical data-augmentation  transformations are limited in expressing more realistic distribution shifts, such as common corruptions~\cite{Hendrycks2019Benchmarking}. More complex operations that are based on compositions of transformations would be preferable~\cite{Hendrycks2020AugMix}.

\begin{figure}[t!]
    \centering
    \includegraphics[width=0.85\columnwidth]{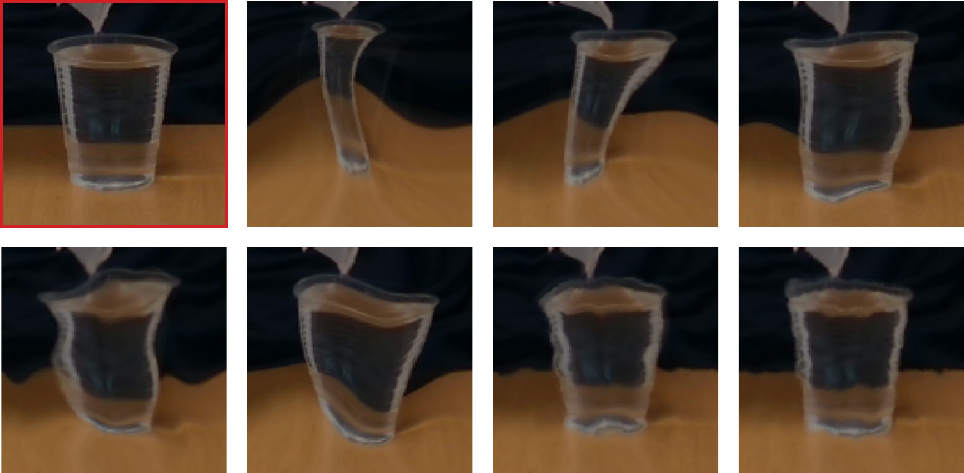}
    \caption{Sample transformed images  using  diffeomorphisms with varying smoothness level  $K_\tau$ (top row, from left to right: original image, transformed image with $K_\tau=2, 5, 10$; bottom row: $K_\tau=20, 40, 100, 300$).}
    \label{fig:diffeo_smoothnes}
    \vspace*{-10pt}
\end{figure}

In this work, we address the filling-level classification training problem using data augmentation with a family of max-entropy transformations, along with a mixing strategy~\cite{Modas2021PRIME}. These transformations allow us to generate diverse and targeted augmentations that can be tuned to generate samples tailored to train classifiers that generalize on containers with shapes (Fig.~\ref{fig:diffeo_smoothnes}), color, and spectral content that are not represented in the original training data. This  constructive approach for augmentation leads to  a filling-level classification accuracy that is on-par with, or better than, the accuracy of transfer learning~\cite{Modas2021Improving}, but with a much smaller training dataset. This augmentation scheme requires only $1.2\times$ additional training time compared to standard training on a small dataset  suitable for estimating the filling level ($\small{\sim}10$ thousand images) and is computationally much less expensive than pre-training on ImageNet ($\small{\sim}1.2$ million images). We also show that the performance of the classifier may further increase when this data augmentation scheme is used in concert with transfer learning itself~\cite{Modas2021Improving}.

%Fig.~\ref{fig:illustration}, PRIME is able to effectively transform the properties of the containers.

%%%%%%%%%%%%%%%%%%%%%%%%%%
\section{Augmentations for filling-level classification}
\label{sec:filling_level_estimation}

We approach the problem of estimating the filling level, $y$, of a container in an image $\x$, as a classification task~\cite{Mottaghi2017ICCV,Modas2021Improving}. We express the filling level as $y \in \{\text{\emph{empty}},\text{\emph{half-full}},\text{\emph{full}},\text{\emph{unknown}}\}$, where the \emph{unknown} class covers cases where the filling level cannot be estimated directly (e.g.~opaque containers). We curated C-CCM~\cite{Modas2021Improving}, a subset of the CORSMAL Containers Manipulation dataset~\cite{Xompero_CCM} with images of cups and glasses that may contain water, pasta or rice\footnote{Note, that, C-CCM contains crops of images around the object of interest. For explicitly transforming only the container in a real human-to-robot application, one might need to first segment the region of interest with the container.}. The shape and transparencies of the  containers, which are captured under different illumination conditions, background types and occlusion levels, vary substantially. We defined three training and validation splits~\cite{Modas2021Improving}, where a distribution shift is introduced in the validation set, such that some containers always have a property that does not exist in the training set (e.g.~a unique shape or color).

%%%%%%%%%%%%%%%%%%%%%%
\subsection{Adversarial transfer learning}
\label{subsec:training_strategies}

With standard training (ST), the classifier learns directly on the available data for the target task. For small datasets, this typically results in classifiers with low accuracy on the test set due to overfitting~\cite{Mottaghi2017ICCV,Modas2021Improving}. Overfitting is caused by the over-reliance of the network on specific, possibly spurious, features that maximize the training accuracy, but do not capture the variability of the categories of interest at test time. 

Transfer learning helps preventing the network from overfitting, by first performing ST on a large dataset representing the \emph{source domain} (e.g.~ImageNet), and then fine-tuning (FT) the resulting network on a smaller dataset representing the \emph{target domain} (e.g.~C-CCM for  filling-level classification). We denote standard transfer learning as ST$\rightarrow$FT. 

The performance of transfer learning for filling-level classification further improves if the network is first adversarially trained in the source domain~\cite{Modas2021Improving}. Adversarial training (AT)~\cite{madryDeepLearningModels2018} is a data augmentation technique that replaces, during training, the original images with their adversarial examples~\cite{szegedyIntriguingPropertiesNeural2014, moosavi-dezfooliDeepFoolSimpleAccurate2016}. We denote this transfer learning approach that uses adversarial learning in the source domain  as AT$\rightarrow$FT. 

% \begin{figure}[t!]
%     \centering
%     \includegraphics[width=0.9\columnwidth]{data_splits.pdf}
%     \caption{The C-CCM~\cite{Modas2021Improving} dataset splits ($\text{S}_1$, $\text{S}_2$, $\text{S}_3$). Black lines indicate containers belonging to the train (validation) set of each split. Parentheses denote the number of images for each type of container.}
%     \label{fig:containers}
%     \vspace*{-5pt}
% \end{figure}

%\subsection{Overfitting and distribution shifts}
%\label{subsec:biases_and_overfitting}

Transfer learning, and especially AT$\rightarrow$FT, improves the filling-level classification performance~\cite{Modas2021Improving}. However, applying adversarial training on a large dataset such as ImageNet is extremely expensive. Moreover, exploiting information from datasets like ImageNet might not always relate to the actual information required by the target task. By transferring this knowledge to the new task, there is an expectation (i.e.~not an explicit design choice) that the target network will learn  more generalizable features. However, there is yet no mechanism for transfer learning to  specify explicitly which features of the training data the target network should be more responsive to. 

To avoid the computational cost of training robust models with transfer learning we investigate how to enrich the training set with data augmentations that effectively increase the variability of data with interpretable image modifications, as discussed in the next section. 

%%%%%%%%%%%%%%%%%%%%%%%%%%
\subsection{Principled data augmentation}
\label{sec:improving_generalization_with_data_augmentation}

The performance on the validation set of C-CMM on the ``shifted'' containers is systematically lower than that on containers that share similarities with those in the training set (overfitting)~\cite{Modas2021Improving}. 
To address this limitation, we consider a data augmentation scheme that generates diverse augmentations using a set of primitive max-entropy transformations on the spatial $\tau$, color, $\gamma$, and spectral, $\omega$, domain~\cite{Modas2021PRIME}. We expect these transformations to  relate to the changes we want to introduce during training: container shape through $\tau$, container color through $\gamma$, and  illumination and texture through $\omega$. Each transformation requires only two control parameters, namely one for the smoothness, $K$, and one for the strength, $\sigma^2$. 

% \begin{figure}[t!]
%     \centering
%     \includegraphics[width=0.88\columnwidth]{illustration.pdf}
%     \caption{Examples of principled augmentations with explicit modification of the shape or color properties of the containers to be used for training.}
%     \label{fig:illustration}
%     \vspace*{-10pt}
% \end{figure}

A transformed image $\bm x_T$ is generated through a convex combination of $n$ basic augmentations (width) consisting of the composition of $m$ of its max-entropy transformations (depth)~\cite{Modas2021PRIME}:
\begin{equation}
    \bm x_T = \sum_{i=1}^n \mu_i g^i_1 \circ\dots\circ g^i_m(\bm x): g\in\{\tau,\gamma,\omega\},
\label{eq:prime}
\end{equation}
where $\bm\mu \sim \mathrm{Dirichlet}$ distribution. The width $n$  specifies the number of transformed instances to be used in the convex combination for synthesizing the transformed image $\bm x_T$, while the depth $m$ specifies how many transformations will be sequentially applied on an image. 

The final image $\hat{\bm x}$ is synthesized as a linear combination of the original image $\bm x$ and the result of the convex combination $\bm x_T$:
\begin{equation}
    \hat{\bm x} = (1-p)\;\bm x + p\;\bm x_T \quad \text{with} \quad p\sim\mathrm{Beta}(\alpha,\beta), 
\label{eq:mixing}
\end{equation}
where the parameters $\alpha$ and $\beta$ control the relative importance of the transformed vs.~original image (when $\alpha>\beta$, the values of the pixels of the transformed $\bm x_T$  are given more importance than the values of the pixels of the original image $\bm x$). 

In the next section, we  identify the  parameter values for  the dataset-specific shifts that arise for each validation set of C-CCM~\cite{Modas2021Improving}.

%%%%%%%%%%%%%%%%%%%%%%%%%%
\section{Experimental evaluation}

We conduct experiments on the C-CCM dataset~\cite{Modas2021Improving} using a ResNet-18~\cite{He2016CVPR_ResNet}. From the C-CCM pre-trained models provided by~\cite{Modas2021Improving}, we evaluate the ones trained with ST (baseline), ST$\rightarrow$FT, and AT$\rightarrow$FT, with the latter currently being the best one for classifying C-CCM. Furthermore, we train a model directly on C-CCM using our Principal Augmentation (PA) scheme and, finally, we also explore the combination of fine-tuning an adversarially trained model~\cite{salman2020adversarially} with PA. We denote this strategy as AT$\rightarrow$PA. We evaluate and compare the different methods on the different splits of C-CCM. All model definitions and training procedures are implemented in PyTorch~\cite{paszkePyTorchImperativeStyle}.

\subsection{Settings and analysis of the effect of the parameter values}

\begin{figure}[t!]
    \centering
    \includegraphics[width=0.9\columnwidth]{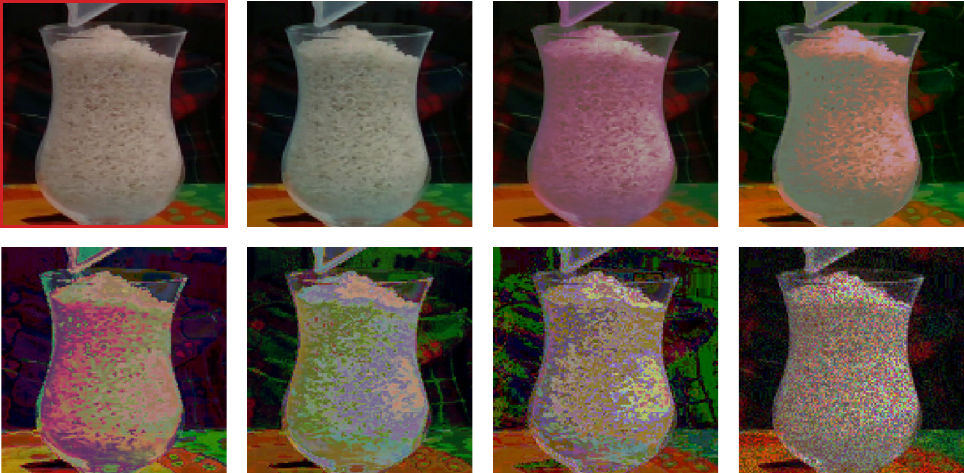}
    \includegraphics[width=0.9\columnwidth]{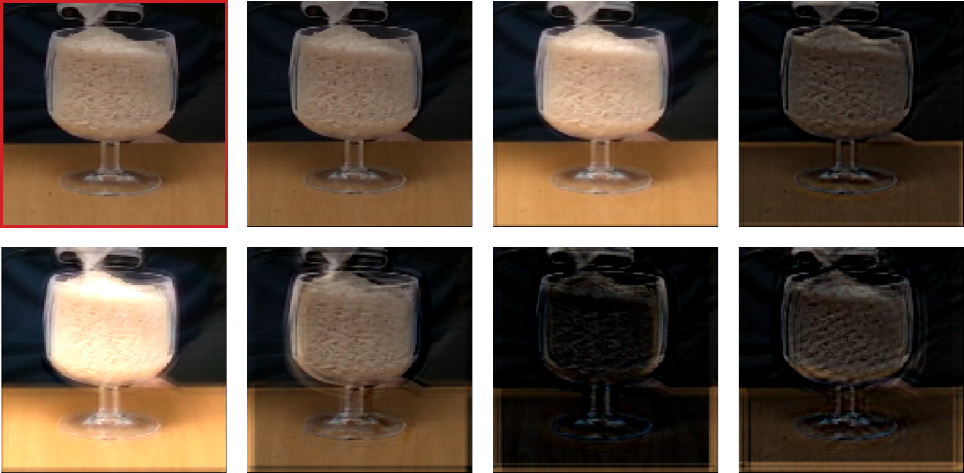}
    \caption{Sample transformed images  using color jittering (top two rows) and spectral filtering (bottom two rows) with varying smoothness level $K_\gamma$ and $K_\omega \times K_\omega$, respectively. First row, from left to right: original image, transformed image with $K_\gamma=2, 5, 10$; second row: $K_\gamma=20, 40, 100, 300$; third row: original image, transformed image with $K_\omega=3, 5, 7$; fourth row: $K_\omega=9, 11, 13, 15$.}
    \label{fig:spectral_colour}
    \vspace*{-10pt}
\end{figure}

\textbf{Split-specific parameters}\quad   The shifts of splits S$_1$ and S$_2$ are mostly related to the shape of the containers. Hence,  we would like to enforce smooth, yet strong, diffeomorphisms that are able to alter the shape of the whole container so it becomes as narrow as a champagne flute, or just a part of it so it resembles the stem of a cocktail glass (see  Fig.~\ref{fig:diffeo_smoothnes}). In fact, for a fixed value of smoothness $K_\tau$ the authors in~\cite{Petrini2021Diffeo} propose to randomly sample the strength $\sigma_\tau^2$ from a specific interval, such that the resulting diffeomorphism remains bijective. In practice, for smaller values of $K_\tau$ (smoother), larger values of $\sigma_\tau^2$ are allowed to be sampled. Hence, we decided to set $K_\tau{=}10$ and let $\sigma_\tau^2$ to be properly sampled during training. In practice we observed that $K\tau\in[10,20]$ still leads to good results. The shifts of split S$_3$ are mostly related to the color and frequency content of the containers (i.e., red and green glass). We therefore focus on the color and spectral transforms (see  Fig.~\ref{fig:spectral_colour}). For the smoothness parameter $K$ of these transformations, we keep the original values~\cite{Modas2021PRIME}: $K_\gamma{=}500$ for the color domain and $K_\omega{=}3$ for the  spectral domain. As for the parameter strength, since very strong changes could remove task-specific information in the images, we only slightly manipulate the color of the pixels and the frequency information of the images, and hence we set $\sigma_\gamma^2{=}0.001$ and $\sigma_\omega^2{=}0.01$, respectively.
%
% \begin{figure}[t!]
%     \centering
%     \includegraphics[width=0.78\columnwidth]{color_smoothness.pdf}
%     \caption{Sample transformed images  using color jittering with varying smoothness level $K_\gamma$ (top row, from left to right: original image, transformed image with $K_\gamma=2, 5, 10$; bottom row: $K_\gamma=20, 40, 100, 300$).}
%     \label{fig:color_smoothnes}
%     \vspace*{-10pt}
% \end{figure}

% \begin{figure}[t!]
%     \centering
%     \includegraphics[width=0.78\columnwidth]{spectral_smoothness.pdf}
%     \caption{Sample transformed images  using  spectral filtering with varying kernel size (smoothness) $K_\omega \times K_\omega$ (top row, from left to right: original image, transformed image with $K_\omega=3, 5, 7$; bottom row: $K_\omega=9, 11, 13, 15$).}
%     \label{fig:spectral_smoothnes}
%     \vspace*{-10pt}
% \end{figure}
%

\begin{table}[!t]
    \centering
    \small
    \caption{Validation accuracy of a ResNet-18 on the C-CCM dataset splits ($\text{S}_1, \text{S}_2, \text{S}_3$), when the composition depth $m$ increases. Here, the transformation width is fixed to $n=3$.}
    \aboverulesep=0ex
    \belowrulesep=0ex
    \begin{tabular}{cccc}
        \rule{0pt}{1.1EM}
        $m$ & $\text{S}_1$ &  $\text{S}_2$ & $\text{S}_3$\\
        \midrule
        $1$ & 82.69 & \textbf{73.42} & 67.91 \\
        \midrule
        $2$ & 83.16 & 70.95 & 65.90 \\
        \midrule
        $3$ & \textbf{84.93} & 68.92 & \textbf{75.03}  \\
        \bottomrule
    \end{tabular}
\vspace*{-0.8em}
\label{tab:depth_analysis}
\end{table}

\begin{figure}[!t]
\centering
\includegraphics[width=\columnwidth]{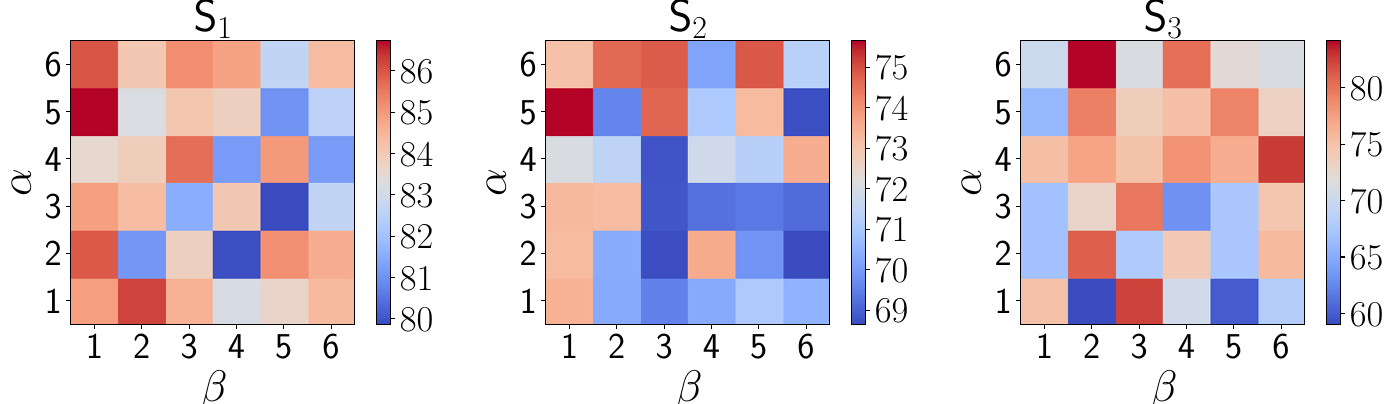}
\caption{Effect of the $\mathrm{Beta}(\alpha, \beta)$ distribution on the validation accuracy of a ResNet-18, trained with our principle augmentation scheme, on each split of C-CCM. Note that, during the mixing step of our scheme, $\alpha>\beta$ imposes more importance to the pixels of the transformed image, while $\alpha<\beta$ to the pixels of the original image.}
\label{fig:sensitivity_beta}
\vspace*{-0.8em}
\end{figure}

%==================================================
% Colors definitions
\definecolor{ts1}{RGB}{ 0  0    0}    % ST
\definecolor{ts2}{RGB}{229, 194, 36}  % AT->FT
\definecolor{ts3}{RGB}{238,113,27}  % PRIME
\definecolor{ts4}{RGB}{19,219,30}  % AT -> PRIME
\definecolor{ts5}{RGB}{0,58,236}  % ST->FT

\pgfplotstableread{accuracy.txt}\accscenarios
\begin{figure*}[!th]
    \vspace{0.8cm}
    \centering
    \begin{tikzpicture}
        \begin{axis}[
            axis x line*=bottom,
            axis y line*=left,
            enlarge x limits=false,
            ybar,
            width=\linewidth,
            bar width=4pt,
            xmin=0,xmax=12,
            xtick=data,
            height=0.4\columnwidth,
            ymin=0,  ymax=100,
            ytick={0,20,40,60,80,100},
            ylabel={Accuracy (\%)},
            % label style={font=\footnotesize},
            tick label style={font=\footnotesize},
            ymajorgrids=true,
            ]
            \addplot+[ybar, black, fill=ts1, draw opacity=0.5] table[x=CoID,y=S1]{\accscenarios};
            \addplot+[ybar, black, fill=ts5, draw opacity=0.5] table[x=CoID,y=S2]{\accscenarios};
            \addplot+[ybar, black, fill=ts2, draw opacity=0.5] table[x=CoID,y=S3]{\accscenarios};
            \addplot+[ybar, black, fill=ts3, draw opacity=0.5] table[x=CoID,y=S4]{\accscenarios};
            \addplot+[ybar, black, fill=ts4, draw opacity=0.5] table[x=CoID,y=S5]{\accscenarios};
        \end{axis}
        \begin{axis}[
            axis x line*=top,
            axis y line*=right,
            width=\linewidth,
            height=.4\columnwidth,
            xmin=0,xmax=12,
            tick label style={
            align=center,
            text width=3cm},
            xtick={2,6,10},
            xticklabels={$\text{S}_1$,$\text{S}_2$,$\text{S}_3$},
            typeset ticklabels with strut,
            % label style={font=\footnotesize},
            ymin=0,ymax=100,
            yticklabels={},
            ]
        \end{axis}
        \node[inner sep=0pt] (flute1) at (-0.5,-0.9)
        {\footnotesize{Validation:}};
        \node[inner sep=0pt] (flute1) at (-0.5,-2.7)
        {\footnotesize{Training:}};
        \node[inner sep=0pt] (flute1) at (1.37,-0.9)
        {\includegraphics[width=.12\columnwidth]{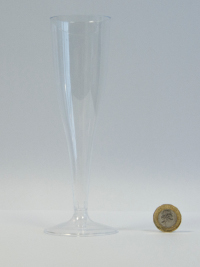}};
         \node[inner sep=0pt] (flute1) at (1.37,-1.8)
        {\scriptsize{(1,059)}};
        \node[inner sep=0pt] (flute1) at (2.75,-0.9)
        {\includegraphics[width=.12\columnwidth]{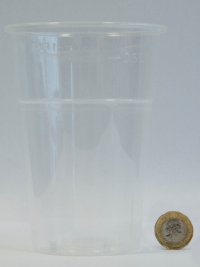}};
        \node[inner sep=0pt] (flute1) at (2.75,-1.8)
        {\scriptsize{(787)}};
        \node[inner sep=0pt] (flute1) at (4.13,-0.9)
        {\includegraphics[width=.12\columnwidth]{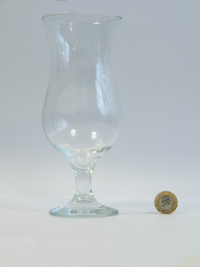}};
        \node[inner sep=0pt] (flute1) at (4.13,-1.8)
        {\scriptsize{(702)}};
        \node[inner sep=0pt] (flute1) at (2.8,-2.7)
        {\includegraphics[width=.53\columnwidth]{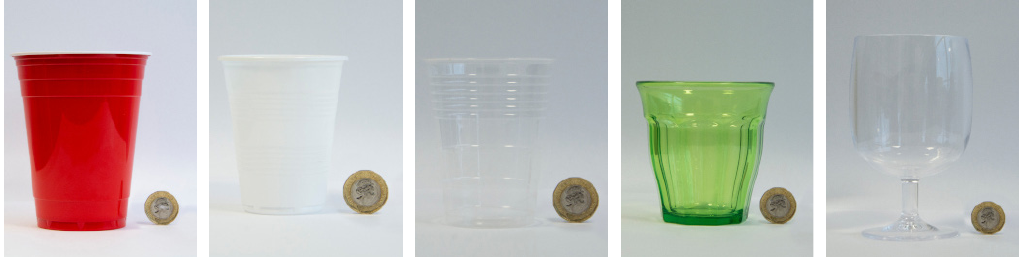}};
        \node[inner sep=0pt] (flute1) at (0.91,-3.49)
        {\scriptsize{(1,623)}};
        \node[inner sep=0pt] (flute1) at (1.87,-3.49)
        {\scriptsize{(1,639)}};
        \node[inner sep=0pt] (flute1) at (2.82,-3.49)
        {\scriptsize{(1,290)}};
        \node[inner sep=0pt] (flute1) at (3.77,-3.49)
        {\scriptsize{(1,566)}};
        \node[inner sep=0pt] (flute1) at (4.71,-3.49)
        {\scriptsize{(1,550)}};
        \node[inner sep=0pt] (flute1) at (6.9,-0.9)
        {\includegraphics[width=.12\columnwidth]{champagne_flute_glass.jpg}};
        \node[inner sep=0pt] (flute1) at (6.9,-1.8)
        {\scriptsize{(1,059)}};
        \node[inner sep=0pt] (flute1) at (8.3,-0.9)
        {\includegraphics[width=.12\columnwidth]{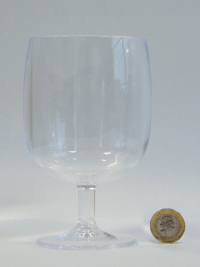}};
        \node[inner sep=0pt] (flute1) at (8.3,-1.8)
        {\scriptsize{(1,550)}};
        \node[inner sep=0pt] (flute1) at (9.69,-0.9)
        {\includegraphics[width=.12\columnwidth]{cocktail_glass.jpg}};
        \node[inner sep=0pt] (flute1) at (9.69,-1.8)
        {\scriptsize{(702)}};
        \node[inner sep=0pt] (flute1) at (8.33,-2.7)
        {\includegraphics[width=.53\columnwidth]{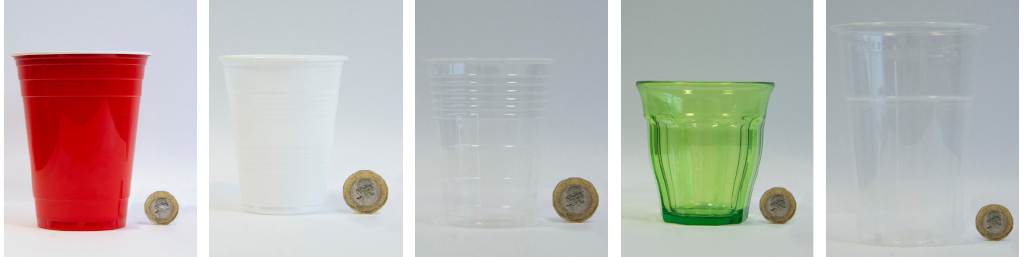}};
        \node[inner sep=0pt] (flute1) at (6.45,-3.49)
        {\scriptsize{(1,623)}};
        \node[inner sep=0pt] (flute1) at (7.4,-3.49)
        {\scriptsize{(1,639)}};
        \node[inner sep=0pt] (flute1) at (8.35,-3.49)
        {\scriptsize{(1,290)}};
        \node[inner sep=0pt] (flute1) at (9.30,-3.49)
        {\scriptsize{(1,566)}};
        \node[inner sep=0pt] (flute1) at (10.21,-3.49)
        {\scriptsize{(787)}};
        \node[inner sep=0pt] (flute1) at (12.45,-0.9)
        {\includegraphics[width=.12\columnwidth]{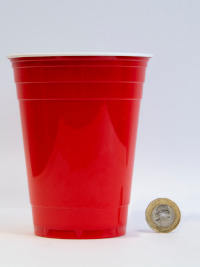}};
        \node[inner sep=0pt] (flute1) at (12.45,-1.8)
        {\scriptsize{(1,624)}};
        \node[inner sep=0pt] (flute1) at (13.81,-0.9)
        {\includegraphics[width=.12\columnwidth]{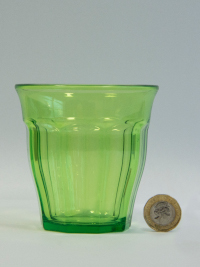}};
        \node[inner sep=0pt] (flute1) at (13.81,-1.8)
        {\scriptsize{(1,566)}};
        \node[inner sep=0pt] (flute1) at (15.2,-0.9)
        {\includegraphics[width=.12\columnwidth]{beer_cup.jpg}};
        \node[inner sep=0pt] (flute1) at (15.2,-1.8)
        {\scriptsize{(787)}};
        \node[inner sep=0pt] (flute1) at (13.9,-2.7)
        {\includegraphics[width=.53\columnwidth]{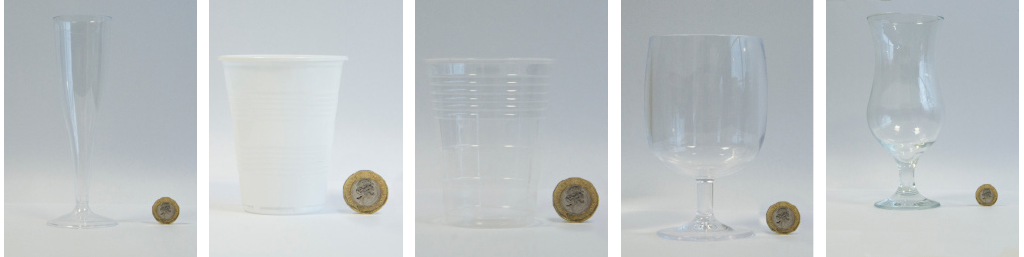}};
        \node[inner sep=0pt] (flute1) at (12.02,-3.49)
        {\scriptsize{(1,059)}};
        \node[inner sep=0pt] (flute1) at (12.96,-3.49)
        {\scriptsize{(1,639)}};
        \node[inner sep=0pt] (flute1) at (13.91,-3.49)
        {\scriptsize{(1,290)}};
        \node[inner sep=0pt] (flute1) at (14.85,-3.49)
        {\scriptsize{(1,550)}};
        \node[inner sep=0pt] (flute1) at (15.79,-3.49)
        {\scriptsize{(702)}};
    \end{tikzpicture}
    \caption{
    Per-container filling level classification accuracy (top) on the three different dataset splits (bottom) of the C-CCM dataset. Parentheses denote the number of images for each type of container. Legend:
    \protect\raisebox{2pt}{\protect\tikz \protect\draw[ts1,line width=2] (0,0) -- (0.3,0);}~ST$^\dagger$,
    \protect\raisebox{2pt}{\protect\tikz \protect\draw[ts5,line width=2] (0,0) -- (0.3,0);}~ST$\rightarrow$FT$^\dagger$.
    \protect\raisebox{2pt}{\protect\tikz \protect\draw[ts2,line width=2] (0,0) -- (0.3,0);}~AT$\rightarrow$FT$^\dagger$,
    \protect\raisebox{2pt}{\protect\tikz \protect\draw[ts3,line width=2] (0,0) -- (0.3,0);}~PA,
    \protect\raisebox{2pt}{\protect\tikz \protect\draw[ts4,line width=2] (0,0) -- (0.3,0);}~AT$\rightarrow$PA. $^\dagger$Values taken from~\protect\cite{Modas2021Improving}.}
    \label{fig:shapeanalysis}
    \vspace{-0.8em}
\end{figure*}
%==================================================

\textbf{Mixing parameters}\quad   It is reasonable to assume that in Eq.~\eqref{eq:prime} we must set $n>1$ in order to increase the diversity of the generated transformed instances. We use the width of the original implementation ($n=3$). In general, it is not always possible to determine the exact outcome of the composition determined by the value of $m$, and its impact on the overall performance.
Hence, we let the mixing coefficient $p$ of Eq.~\eqref{eq:mixing} to be uniformly sampled ($\alpha=\beta=1$) and perform a sensitivity analysis on the values of $m$. The performance of a ResNet-18~\cite{He2016CVPR_ResNet} on each dataset split is shown in Tab.~\ref{tab:depth_analysis}: while for S$_1$ and S$_3$, increasing the depth significantly improves the performance, the opposite happens for S$_2$, suggesting that that applying multiple transformations on the image degrades some important information. Then, for the best values of $m$ in Tab.~\ref{tab:depth_analysis}, we explore the effect of the mixing coefficient $p$ of equation Eq.~\eqref{eq:mixing}. We focus on the parameters of the $\mathrm{Beta}$ distribution, which control the relative importance of the pixels of $\bm x$ or $\bm x_T$. Since the classifier overfits to the training data we would expect that more importance on $\bm x_T$ might be necessary. To that end, we perform a sensitivity analysis on the values of $\alpha$ and $\beta$, by measuring the performance of the network on their different combinations. The results are shown in Fig.~\ref{fig:sensitivity_beta}: on every dataset split, the highest validation accuracy is achieved when more importance is given to the pixels of the transformed image. In particular, on S$_1$ and S$_2$, the best performance ($86.73\%$ and $75.66\%$ respectively) is achieved for $\mathrm{Beta}(5, 1)$, while on S$_3$ ($84.21\%$) it is achieved for $\mathrm{Beta}(6, 2)$.

\textbf{Training and validation}\quad
For PA and AT$\rightarrow$PA strategies we train or fine-tune the classifier for $50$ epochs, using a  cross-entropy loss~\cite{Shalev_Shwartz} and stochastic gradient descent. The maximum learning rate for updating the weights is set to $0.05$ and $0.005$ when performing PA and AT$\rightarrow$PA respectively. The learning rate decays linearly during training. Note that the models we evaluate are the ones that achieve the highest validation accuracy (early-stopping). For dealing with class imbalances, the training images in a batch are randomly sampled with probabilities that are inversely proportional to the number of images of each class. For the case of AT$\rightarrow$PA, since there are multiple source models adversarially trained with perturbations of different strength $\epsilon$, we decided to choose those that lead to the highest validation accuracy. Hence, for S$_1$ we select a network trained with $\epsilon=0.05$, while for S$_2$ and S$_3$ a network trained with $\epsilon=0.5$. Recall that, for the AT$\rightarrow$FT models used in~\cite{Modas2021Improving} the selected values of $\epsilon$ were $0.05$, $1$ and $0.5$ for each dataset split respectively.

%%%%%%%%%%%%%%%%%%%%
\subsection{Classification results and discussion}
\label{subsec:results}

Fig.~\ref{fig:shapeanalysis} shows the classification performance of different strategies on the three configurations, $\text{S}_1$,  $\text{S}_2$ and  $\text{S}_3$. The results indicate that pre-training might not be necessary, since properly tuning data augmentation to compensate for the dataset-specific distribution shifts can improve performance, while requiring a lower computational cost than using transfer learning. PA requires only $1.2\times$ additional training time compared to ST on C-CCM, which is many orders of magnitude lower than (adversarially) training a model on ImageNet for using transfer learning. When training time is not an issue, transfer learning with AT at the source domain combined with PA (AT$\rightarrow$PA) generally improves performance. Note that when the performance of ST is low, all strategies lead to significant improvements; whereas  when ST performs well,  AT$\rightarrow$FT has an  insignificant contribution or decreases the final performance. 

For  $\text{S}_1$, the low performance of ST on the champagne flute (left) is improved by both AT$\rightarrow$FT and PA, and even more so by  AT$\rightarrow$PA, suggesting that diffeomorphisms compensate for the unique narrow shape of the flute.
The accuracy of ST on the beer cup (middle) is high, due to the  shape similarity  of the data in the training set (e.g.~small transparent cup).  AT$\rightarrow$FT causes a small accuracy drop, whereas PA retains the performance and AT$\rightarrow$PA improves it. The accuracy of ST on the cocktail glass (right) is  slightly improved with AT$\rightarrow$FT and considerably improved by PA and AT$\rightarrow$PA. Although there is another container with a stem in the training set (wine glass), it seems that the introduced diffeomorphisms  better compensate for the different shape above the stem of the cocktail glass. 

For  $\text{S}_2$, the accuracy of all strategies on the champagne flute (left) and the cocktail glass (right) is somehow similar in trend  to that on $\text{S}_1$. Note that there are no containers with a stem in the training set. Yet, the performance on the wine glass (middle) is similar for most strategies, which might be due to the similarity of its shape above the stem with the other transparent cups in the training set.

For  $\text{S}_3$, there is no colored container in the training set. ST is unable to generalize for the red cup (left), unlike AT$\rightarrow$FT, PA and AT$\rightarrow$PA. Still, the accuracy with data augmentation is not on the same level as with AT$\rightarrow$FT, which sets this specific container case as an example of the benefits of adversarially pre-training the network on a large and diverse source dataset.
As for the green glass (middle) AT$\rightarrow$FT increases on ST, similarly to  AT$\rightarrow$PA.
Finally, the accuracy of ST on the beer cup (right) is high and the other strategies cannot reach that level, with AT$\rightarrow$PA featuring the lowest performance drop. 

%%%%%%%%%%%%%%%%%%%%%%%%%%%%%%%%%%%%%%%%%%%%%%%%%%
\section{Conclusion}
\label{sec:conclusion}

We investigated the impact of data augmentation on classifying the filling level of a container, when the containers in the test-time images have different properties than those in the training set. We compared transfer learning -- with or without adversarial training~\cite{Modas2021Improving} -- and a principled augmentation approach that explicitly operates on the geometry, color and spatial frequencies of the training images~\cite{Modas2021PRIME} in order to generalize the shape, color, and spectral content of an available, limited training dataset. We showed that the principled augmentation can either replace transfer learning approaches, which are computationally more expensive, or be combined with adversarial transfer learning to improve its performance. 

As future work, we will incorporate new transformations to compensate for other types of distribution shifts related to transparencies and occlusions. Furthermore, we will extend our analysis to other datasets and settings.

\section*{Acknowledgments}
This work is supported by the CHIST-ERA program through the project CORSMAL, under UK EPSRC grant EP/S031715/1 and Swiss NSF grant 20CH21{\_}180444.

\bibliographystyle{IEEEtran}
\bibliography{strings,refs}

\begin{thebibliography}{10}
\providecommand{\url}[1]{#1}
\csname url@samestyle\endcsname
\providecommand{\newblock}{\relax}
\providecommand{\bibinfo}[2]{#2}
\providecommand{\BIBentrySTDinterwordspacing}{\spaceskip=0pt\relax}
\providecommand{\BIBentryALTinterwordstretchfactor}{4}
\providecommand{\BIBentryALTinterwordspacing}{\spaceskip=\fontdimen2\font plus
\BIBentryALTinterwordstretchfactor\fontdimen3\font minus
  \fontdimen4\font\relax}
\providecommand{\BIBforeignlanguage}[2]{{%
\expandafter\ifx\csname l@#1\endcsname\relax
\typeout{** WARNING: IEEEtran.bst: No hyphenation pattern has been}%
\typeout{** loaded for the language `#1'. Using the pattern for}%
\typeout{** the default language instead.}%
\else
\language=\csname l@#1\endcsname
\fi
#2}}
\providecommand{\BIBdecl}{\relax}
\BIBdecl

\bibitem{TanTransferLearning}
C.~Tan, F.~Sun, T.~Kong, W.~Zhang, C.~Yang, and C.~Liu, ``A survey on deep
  transfer learning,'' in \emph{Int. Conf. on Artificial Neural Networks},
  2018, pp. 270--279.

\bibitem{huh2016makes}
M.~Huh, P.~Agrawal, and A.~A. Efros, ``What makes {ImageNet} good for transfer
  learning?'' in \emph{Neural Inf. Process. Syst. Workshop on Large Scale
  Computer Vision Systems}, Dec. 2016.

\bibitem{Kornblith_2019_CVPR}
S.~Kornblith, J.~Shlens, and Q.~V. Le, ``Do better {ImageNet} models transfer
  better?'' in \emph{Proc. IEEE Conf. Comput. Vis. Pattern Recognit.}, Jun.
  2019.

\bibitem{XueTexture}
J.~{Xue}, H.~{Zhang}, and K.~{Dana}, ``Deep texture manifold for ground terrain
  recognition,'' in \emph{Proc. IEEE Conf. Comput. Vis. Pattern Recognit.},
  2018.

\bibitem{Farhadi2009}
A.~{Farhadi}, I.~{Endres}, D.~{Hoiem}, and D.~{Forsyth}, ``Describing objects
  by their attributes,'' in \emph{Proc. IEEE Conf. Comput. Vis. Pattern
  Recognit.}, Jun. 2009.

\bibitem{Sajjan2020ICRA_ClearGrasp}
S.~S. Sajjan, M.~Moore, M.~Pan, G.~Nagaraja, J.~Lee, A.~Zeng, and S.~Song,
  ``{ClearGrasp}: {3D} shape estimation of transparent objects for
  manipulation,'' in \emph{Proc. IEEE Int. Conf. Robotics Autom.}, Jun. 2020.

\bibitem{Vedaldi2014CVPR}
A.~Vedaldi, S.~Mahendran, S.~Tsogkas, S.~Maji, R.~Girshick, J.~Kannala,
  E.~Rahtu, I.~Kokkinos, M.~B. Blaschko, D.~Weiss, B.~Taskar, K.~Simonyan,
  N.~Saphra, and S.~Mohamed, ``Understanding objects in detail with
  fine-grained attributes,'' in \emph{Proc. IEEE Conf. Comput. Vis. Pattern
  Recognit.}, Jun. 2014.

\bibitem{Shorten2019SurveyDataAugm}
C.~Shorten and T.~M. Khoshgoftaar, ``A survey on image data augmentation for
  deep learning,'' \emph{Journal of Big Data}, vol.~6, no.~1, 2019.

\bibitem{Hendrycks2020AugMix}
D.~Hendrycks, N.~Mu, E.~D. Cubuk, B.~Zoph, J.~Gilmer, and B.~Lakshminarayanan,
  ``{AugMix}: A simple method to improve robustness and uncertainty under data
  shift,'' in \emph{Proc. Int. Conf. Learning Represent.}, Apr. 2020.

\bibitem{Sanchez-Matilla2020}
R.~{Sanchez-Matilla}, K.~{Chatzilygeroudis}, A.~{Modas}, N.~{Ferreira Duarte},
  A.~{Xompero}, P.~{Frossard}, A.~{Billard}, and A.~{Cavallaro}, ``Benchmark
  for human-to-robot handovers of unseen containers with unknown filling,''
  \emph{IEEE Robotics Autom. Lett.}, vol.~5, no.~2, Apr. 2020.

\bibitem{Xompero2020ICASSP_LoDE}
A.~Xompero, R.~Sanchez-Matilla, A.~Modas, P.~Frossard, and A.~Cavallaro,
  ``Multi-view shape estimation of transparent containers,'' in \emph{Proc.
  IEEE Int. Conf. Acoustics, Speech Signal Process.}, May 2020.

\bibitem{Mottaghi2017ICCV}
R.~Mottaghi, C.~Schenck, D.~Fox, and A.~Farhadi, ``See the glass half full:
  Reasoning about liquid containers, their volume and content,'' in \emph{Proc.
  IEEE Int. Conf. Comput. Vis.}, Oct 2017.

\bibitem{Modas2021Improving}
A.~Modas, A.~Xompero, R.~{Sanchez-Matilla}, P.~Frossard, and A.~Cavallaro,
  ``Improving filling level classification with adversarial training,'' in
  \emph{Proc. IEEE Int. Conf. Image Process.}, 2021, pp. 829--833.

\bibitem{Schenck2017ICRA}
C.~Schenck and D.~Fox, ``Visual closed-loop control for pouring liquids,'' in
  \emph{Proc. IEEE Int. Conf. Robotics Autom.}, May 2017.

\bibitem{Do2016}
C.~Do, T.~Schubert, and W.~Burgard, ``A probabilistic approach to liquid level
  detection in cups using an {RGB-D} camera,'' in \emph{Proc. IEEE Int. Conf.
  Intell. Robot Syst.}, Oct. 2016.

\bibitem{Do2018}
C.~Do and W.~Burgard, ``Accurate pouring with an autonomous robot using an
  {RGB-D} camera,'' in \emph{Proc. Int. Conf. Intell. Auton. Syst.}, Jul. 2018.

\bibitem{Deng2009CVPR_ImageNet}
J.~Deng, W.~Dong, R.~Socher, L.-J. Li, L.~Li, and L.~Fei-Fei, ``{ImageNet}: A
  large-scale hierarchical image database,'' in \emph{Proc. IEEE Conf. Comput.
  Vis. Pattern Recognit.}, Jun. 2009.

\bibitem{Hendrycks2019Benchmarking}
D.~Hendrycks and T.~Dietterich, ``Benchmarking neural network robustness to
  common corruptions and perturbations,'' in \emph{Proc. Int. Conf. Learning
  Represent.}, May 2019.

\bibitem{Modas2021PRIME}
A.~Modas, R.~Rade, G.~{Ortiz-Jim\'enez}, S.-M. {Moosavi-Dezfooli}, and
  P.~Frossard, ``{PRIME}: {A} few primitives can boost robustness to common
  corruptions,'' \emph{arXiv:2112.13547}, Dec. 2021.

\bibitem{Xompero_CCM}
\BIBentryALTinterwordspacing
A.~Xompero, R.~Sanchez-Matilla, R.~Mazzon, and A.~Cavallaro, ``{CORSMAL
  Containers Manipulation},'' 2020, (1.0) [Dataset]. Queen Mary University of
  London. \url{https://doi.org/10.17636/101CORSMAL1}. [Online]. Available:
  \url{http://corsmal.eecs.qmul.ac.uk/containers_manip.html}
\BIBentrySTDinterwordspacing

\bibitem{madryDeepLearningModels2018}
A.~Madry, A.~Makelov, L.~Schmidt, D.~Tsipras, and A.~Vladu, ``Towards deep
  learning models resistant to adversarial attacks,'' in \emph{Proc. Int. Conf.
  Learning Represent.}, Apr. 2018.

\bibitem{szegedyIntriguingPropertiesNeural2014}
C.~Szegedy, W.~Zaremba, I.~Sutskever, J.~Bruna, D.~Erhan, I.~Goodfellow, and
  R.~Fergus, ``Intriguing properties of neural networks,'' in \emph{Proc. Int.
  Conf. Learning Represent.}, Apr. 2014.

\bibitem{moosavi-dezfooliDeepFoolSimpleAccurate2016}
S.~{Moosavi-Dezfooli}, A.~Fawzi, and P.~Frossard, ``{{DeepFool}}: {{A simple}}
  and accurate method to fool deep neural networks,'' in \emph{Proc. IEEE Conf.
  Comput. Vis. Pattern Recognit.}, Jun. 2016.

\bibitem{He2016CVPR_ResNet}
K.~He, X.~Zhang, S.~Ren, and J.~Sun, ``Deep residual learning for image
  recognition,'' in \emph{Proc. IEEE Conf. Comput. Vis. Pattern Recognit.},
  Jun. 2016.

\bibitem{salman2020adversarially}
H.~Salman, A.~Ilyas, L.~Engstrom, A.~Kapoor, and A.~Madry, ``Do adversarially
  robust {ImageNet} models transfer better?'' in \emph{Adv. Neural Inf.
  Process. Syst.}, Dec. 2020.

\bibitem{paszkePyTorchImperativeStyle}
A.~Paszke, S.~Gross, F.~Massa, A.~Lerer, J.~Bradbury, G.~Chanan, T.~Killeen,
  Z.~Lin, N.~Gimelshein, L.~Antiga, A.~Desmaison, A.~Kopf, E.~Yang, Z.~DeVito,
  M.~Raison, A.~Tejani, S.~Chilamkurthy, B.~Steiner, L.~Fang, J.~Bai, and
  S.~Chintala, ``{{PyTorch}}: {{An imperative style}}, {{high}}-{{performance
  deep learning library}},'' in \emph{Adv. Neural Inf. Process. Syst.}, Dec.
  2019.

\bibitem{Petrini2021Diffeo}
L.~Petrini, A.~Favero, M.~Geiger, and M.~Wyart, ``Relative stability toward
  diffeomorphisms indicates performance in deep nets,'' in \emph{Adv. Neural
  Inf. Process. Syst.}, Dec. 2021.

\bibitem{Shalev_Shwartz}
S.~{Shalev-Shwartz} and S.~{Ben-David}, \emph{Understanding Machine Learning:
  From Theory to Algorithms}.\hskip 1em plus 0.5em minus 0.4em\relax Cambridge
  University Press, 2014.

\end{thebibliography}

\end{document}